\documentclass[journal]{IEEEtai}

\usepackage[colorlinks,urlcolor=blue,linkcolor=blue,citecolor=blue]{hyperref}

\usepackage{color,array,amsmath,bm}
\usepackage{algorithm}
\usepackage{algorithmic}
\usepackage{graphicx}
\usepackage{overpic}
\usepackage{soul} 
\usepackage{graphicx}
\usepackage{multirow}
\usepackage{booktabs}
\usepackage{cprotect}  
\usepackage{newfloat}
\usepackage{listings}
\usepackage{amssymb}
\usepackage[a4paper, total={7in, 10in}]{geometry}
\usepackage[font=footnotesize]{caption}

\begin{document}


\title{Evolving Efficient Genetic Encoding for Deep Spiking Neural Networks} 
\author{Wenxuan Pan, Feifei Zhao, Bing Han, Haibo Tong, Yi Zeng

\thanks{Wenxuan Pan is with the Brain-inspired Cognitive Intelligence Lab, Institute of Automation, Chinese Academy of Sciences, Beijing 100190, China, and School of Artificial Intelligence, University of Chinese Academy of Sciences, Beijing 100049, China.} 
\thanks{Feifei Zhao is with the Brain-inspired Cognitive Intelligence Lab, Institute of Automation, Chinese Academy of Sciences, Beijing 100190, China.} 
\thanks{Bing Han is with the Brain-inspired Cognitive Intelligence Lab, Institute of Automation, Chinese Academy of Sciences, Beijing 100190, China, and School of Artificial Intelligence, University of Chinese Academy of Sciences, Beijing 100049, China.} 
\thanks{Haibo Tong is with the Brain-inspired Cognitive Intelligence Lab, Institute of Automation, Chinese Academy of Sciences, Beijing 100190, China, and School of Artificial Intelligence, University of Chinese Academy of Sciences, Beijing 100049, China.} 
\thanks{Yi Zeng is with the Brain-inspired Cognitive Intelligence Lab, Institute of Automation, Chinese Academy of Sciences, Beijing 100190, China, and Center for Long-term Artificial Intelligence, Beijing 100190, China, and University of Chinese Academy of Sciences, Beijing 100083, China, and Key Laboratory of Brain Cognition and Brain-inspired Intelligence Technology, Chinese Academy of Sciences, Shanghai, 200031, China.}

\thanks{The first and the second authors contributed equally to this work, and serve as co-first authors.}
\thanks{The corresponding author is Yi Zeng (e-mail: yi.zeng@ia.ac.cn).}}

\maketitle

\begin{abstract}
By exploiting discrete signal processing and simulating brain neuron communication, Spiking Neural Networks (SNNs) offer a low-energy alternative to Artificial Neural Networks (ANNs). However, existing SNN models, still face high computational costs due to the numerous time steps as well as network depth and scale. The tens of billions of neurons and trillions of synapses in the human brain are developed from only 20,000 genes, which inspires us to design an efficient genetic encoding strategy that dynamic evolves to regulate large-scale deep SNNs at low cost. Therefore, we first propose a genetically scaled SNN encoding scheme that incorporates globally shared genetic interactions to indirectly optimize neuronal encoding instead of weight, which obviously brings about reductions in parameters and energy consumption. Then, a spatio-temporal evolutionary framework is designed to optimize the inherently initial wiring rules. Two dynamic regularization operators in the fitness function evolve the neuronal encoding to a suitable distribution and enhance information quality of the genetic interaction respectively, substantially accelerating evolutionary speed and improving efficiency. Experiments show that our approach compresses parameters by approximately 50\% to 80\%, while outperforming models on the same architectures by 0.21\% to 4.38\% on CIFAR-10, CIFAR-100 and ImageNet. In summary, the consistent trends of the proposed genetically encoded spatio-temporal evolution across different datasets and architectures highlight its significant enhancements in terms of efficiency, broad scalability and robustness, demonstrating the advantages of the brain-inspired evolutionary genetic coding for SNN optimization.

\end{abstract}



\section{Introduction}
Artificial neural networks have provided important insights into numerous application areas~\cite{he2016deep,mikolov2013efficient}, but the large number of matrix operations increases significantly with the size of the network.
Spiking Neural Networks~\cite{maass1997networks}, as the third generation of neural networks, achieve a low-energy computing paradigm by simulating the communication characteristics of brain neurons and leveraging the inherent energy efficiency advantages of discrete signal processing and are more biologically plausible.
Most of the research on SNN optimization focuses on the training mechanism, which can be roughly  divided into plasticity-based training~\cite{diehl2015fast}, conversion-based training~\cite{han2020rmp}, and gradient-based training~\cite{wu2018spatio}, which has become the most mainstream method due to its high efficiency. However, a large number of time steps, proxy gradient calculations, and increasingly deeper network architecture designs also make SNNs increasingly computationally expensive. 
Common techniques for optimizing computational overhead include network pruning and quantization~\cite{shen2024conventional}, which greatly reduce storage and computational costs by pruning redundant connections~\cite{han2023enhancing} and reducing operations~\cite{nagel2020up}.
Despite offering numerous insights and successful methodologies for training on complex tasks, current SNN models still lack effective integration with brain-inspired mechanisms to achieve a balance between cost and performance.

Generation after generation, environmental pressures drive biological neural systems to evolve specific responses to complex tasks, with neural connections continually adapting to encode crucial information into the connectome~\cite{barabasi2023complex}.
Evolution has endowed approximately 21,000 genes with the ability to support the complex computing capabilities of the brain's $10^{10}$ neurons and $10^{15}$ synapses.
This compact and efficient encoding method not only saves biological energy, but also facilitates genetic optimization during evolution, thereby supporting complex cognitive functions and highly flexible behavioral adaptability.
This inspired us to design a gene-scaled neuronal coding paradigm for SNNs, controlling the entire network with very few parameters,
and simulate the evolutionary process of the brain to optimize the genetic encoding.

Based on this, this paper implements genetically encoded neural networks on a variety of common network architectures, re-encoding weights with neuronal encoding each layer and global shared gene interaction matrices, greatly compressing parameters and energy consumption.
Furthermore, to accelerate the evolution and training, we propose a spatio-temporal evolutionary SNN framework. 
Spatially, the initial wiring of neuronal encodings and gene interaction matrix are optimized through the Covariance Matrix Adaptation Evolution Strategy (CMA-ES).
Temporally, the dynamic regularization scale helps solutions transition from freely learning complex features to gradually establishing stable functional patterns, ultimately achieving a significant reduction in energy consumption without compromising performance. 
By integrating principles of neuroevolution, the proposed method develops robust, efficient architectures capable of performing complex tasks at very low computational cost, reflecting the evolutionary encoding of behaviors in the brain's connectome. In general, the contributions of this work can be summarized as follows:

\begin{itemize}

\item[1)] We develop a Genetically Encoded Evolutionary (GEE) spiking neural network that improves the performance by learning neuronal encoding rather than directly updating weights. The controllable genetic scale greatly reducing the number of parameters and computational cost without losing accuracy. 
\item[2)] To improve the quality of solutions, we propose a Spatio-Temporal dynamical Evolution (STE) framework for initial wiring of neuronal encoding and gene interactions. 
Two dynamic regularization operators, spatial entropy and temporal difference regularization, help improve the evolution efficiency and greatly reduce the cost.
\item[3)] On CIFAR10, CIFAR100 and ImageNet datasets, the proposed GEE achieves superior performance with significantly lower energy consumption, achieving efficient SNN evolution with a brain-inspired efficient computational paradigm.

\end{itemize}

\section{Related Work}
This paper proposes a brain-inspired genetically encoded spiking neural network, supplemented by a spatio-temporal evolution framework to optimize the initial wiring. To this end, we first analyze energy-efficient network design techniques, and then discuss related work on tensor decomposition methods and natural evolution strategies, respectively.
\subsection{Efficient Network Designs}
The discrete spiking communication of SNN simulates the behavior of biological neurons, providing a more energy-efficient alternative to ANN. 
Optimization of computational cost has attracted research attention and has become an important trajectory in promoting the development of deep learning.
Structurally, the work of pruning SNN structures~\cite{han2023enhancing} not only significantly reduces the computational and storage costs by pruning redundant neurons, but also maintained the performance comparable to the ANN models.
On the other hand, quantization-based methods optimize weights by reducing floating-point multiplications~\cite{lin2015neural,nagel2020up}, reducing the amount of calculation consumption and balancing performance~\cite{shen2024conventional}.
From a brain-inspired perspective, this work designs a gene-scaled neuronal coding paradigm that simultaneously optimizes wiring and weights to significantly reduce computational cost without losing accuracy.


\subsection{Tensor Decomposition Methods}
Although not inspired by this, the genetic encoding adopted in this work is mathematically similar to a kind of tensor decomposition methods~\cite{kolda2009tensor}, which decomposes a high-dimensional tensor into the product of several low-dimensional tensors to help extract features and simplify data structures. According to the different decomposition methods, it can be divided into CP decomposition, Tucker decomposition and Tensor-Train (TT) decomposition, etc.
Its significant advantage is that it changes the way to optimize and enhance the network by reparameterization, and has achieved significant success in many fields of deep learning~\cite{101109,Factorized,dengTensor,liu2023tensor}.

There is not much work on applying tensor decomposition to optimize SNNs. \cite{lee2024tt} applies parallel TT decomposition to decompose the SNN convolutional layer weights into smaller tensors, reducing the number of parameters, FLOPs and training cost, but at the expense of some accuracy.
\cite{dengTensor} introduces tensor decomposition into the attention module and uses CP decomposition to dynamically adjust the rank of the attention map in SNN according to the task, but it does not involve the structure or encoding of the entire network.
This paper is inspired by the genetic connection mechanism of the brain's neural circuits (rather than the tensor decomposition method) which is more consistent with the genetic mechanism of the nervous system.


\subsection{Natural Evolution Strategies}
Natural Evolution Strategies (NES)~\cite{wierstra2014natural} is a family of optimization algorithms that employ gradient-based updates to evolve a population of candidate solutions towards better performance on a given task. 
NES optimizes the parameters of a probability distribution to minimize the expected evaluation of solutions, rather than directly seeking the optimal solution to an objective function~\cite{nomura2022fast}, making it well suited for continuous parameter optimization and fast convergence. This inspired us to adopt NES to optimize the relevant parameters of the proposed genetic encoding network, promoting the improvement of evolution efficiency and solution quality.


\section{Method}

\begin{figure*}[htp]
\centering
\includegraphics[width=17.5cm]{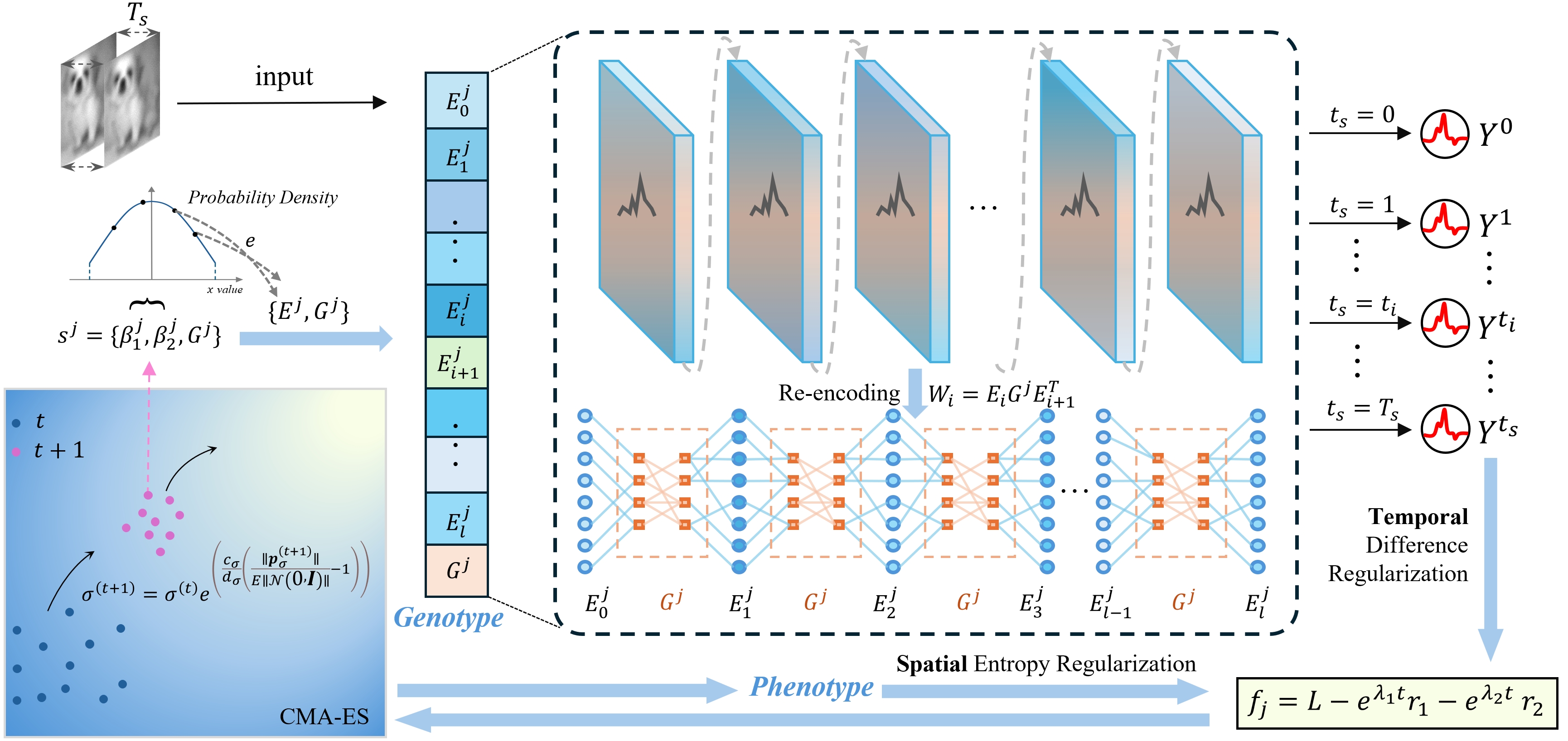}
\caption{Spatio-temporal evolving genetically encoded spiking neural networks.}
\label{method}
\end{figure*}

\subsection{Indirectly Genetic Encoding for Deep SNNs}
The workings of neural systems in the human brain are influenced by neuronal connections, learning, and random noise. Existing works have abstracted these mechanisms as functions of the interaction between neuronal identity and genetic factors~\cite{barabasi2020genetic,barabasi2023complex}.
Based on this principle, we reconstruct deep spiking neural networks into genetically encoded efficient SNNs based on the Leaky Integrate and Fire (LIF) neuron model~\cite{lapicque1907recherches}.
The framework is shown as Fig.~\ref{method}.
Assume the input channels are $C_{in}$, the output channels are $C_{out}$, and the kernel size is $k \times k$. The weights of a conventional convolutional layer $W \in \mathbb{R}^{C_{out} \times C_{in} \times k \times k}$. In our genetic convolutional layer, the weights are constructed based on the indirectly genetic encoding of input neurons $E_1$ and output neurons $E_2$ and can be expressed as:
\begin{equation}
W=E_1GE_2^T
\end{equation}
where $E_1\in \mathbb{R}^{C_{in}\times k\times k\times g}$, $E_2 \in \mathbb{R}^{C_{out} \times k \times k \times g}$. $G$ is the gene interaction matrix, $G \in \mathbb{R}^{g \times g}$. 
Such genetically encoding method greatly compresses the overall parameters of the network, from the original weight size $C_{out} * C_{in} *  k *  k$, to re-encoded parameters determined by a small genes $g$:
\begin{equation}
Parameters = g\left(C_{in}k^2 + g + C_{out}k^2\right)
\end{equation}
In a deep network, except for the first and last layers, each layer acts both as the output of the previous layer and the input of the next layer. 
Therefore, the weight of the $i_{th}$ layer is encoded by the neuronal encoding $E_i$ of this layer, the neuronal encoding $E_{i+1}$ of the next layer, and the gene interaction matrix $G$:
\begin{equation}
W_i=E_{i}GE_{i+1}^T
\label{w}
\end{equation}
To calculate the gradient of the loss \(L\) with respect to \(E_i\), we need to apply the chain rule during backpropagation.
Assume that the output of each layer is expressed as $X_{i+1} = f(W_i X_i) $, where \(f\) is the activation function. Let $\delta_{i}$ represent the gradient of the loss with respect to the output of $i_{th}$ layer. 
The gradient of the weights with respect to the parameters \(E_i\) is $G E_{i+1}^T$ and the gradient of the output of layer \(i+1\) with respect to the weights is $ f'(W_i X_i)  X_i^T$,
where \(f\) is the activation function and \(X_i\) is the input to the $i_{th}$ layer. Therefore, for the last layer, the gradient of the weights with respect to \(E_{i+1}\) is $ G^T E_{i}^T$. Therefore, the gradient is updated as:
\begin{equation}
E_{i} \leftarrow E_{i} - \eta \left( \delta_{i} f'(E_{i-1}GE_i^T X_{i-1})  X_{i-1}^T (G^T E_{i-1}^T) \right)
\end{equation}
where \(\eta\) is the learning rate. 
For all previous layers, the parameters \(E_i\) can be updated using gradient descent as follows: 
\begin{equation}
E_i \leftarrow E_i - \eta  \left( \delta_{i+1}  f'(E_i G E_{i+1}^T X_i)  X_i^T  (G  E_{i+1}^T)\right)
\end{equation}


Theoretically, any $m \times n$ matrix $W$ can be expressed through Singular Value Decomposition (SVD) as:
\begin{equation}
W = U \Sigma V^T
\end{equation}
where $U$ is an $m \times m$ orthogonal matrix, $V$ is an $n \times n$ orthogonal matrix, and $\Sigma$ is an $m \times n$ diagonal matrix, with the diagonal elements being the singular values. If the rank of $W$ is $r_w$, then $\Sigma$ contains only $r_w$ non-zero singular values. For a low-rank approximation, one can select the top $r'$ largest singular values, where $r' \leq r_w$, and construct the approximate matrix:
\begin{equation}
W' = U' \Sigma' V'^T
\end{equation}
where $U'$ and $V'$ are the columns of $U$ and $V$ corresponding to the largest $r'$ singular values, and $\Sigma'$ is a $r' \times r'$ diagonal matrix. If we define $E_1$ as $U' \Sigma'^{1/2}$, $E_2$ as $\Sigma'^{1/2} V'^T$, and $G$ as any $r' \times r'$ matrix that encapsulates gene interactions, then:
\begin{equation}
E_1 G E_2 = (U' \Sigma'^{1/2}) G (\Sigma'^{1/2} V'^T) = U' (\Sigma'^{1/2} G \Sigma'^{1/2}) V'^T
\end{equation}
Thus, the gene encoding $E_1 G E_2$ provides a way to incorporate gene interactions encoded in $G$ into the low-rank approximation of $W$. 
The weights approximated in this way are able to capture more complex behaviors or structures, since $G$ allows the approximate representation of $W$ to be enriched and adjusted in a low-rank format.
\subsection{Spatial-temporal Dynamic Evolution}

As the network becomes deeper, multiple matrix multiplications and indirect updates will lead to slow training progress or even vanishing gradients. However, it is very difficult to optimize all neuronal encodings because of the high dimensionality.
Therefore, we choose to indirectly control the solution by evolving the distribution of the initial $E_i^j$ and the global-shared gene interactions $G^j$, which significantly reduces the number of parameters and improves the efficiency of evolution. 
Each phenotype $\{E^j_0, E^j_1... E^j_l,G^j\}$ represents an architecture, including the gene interactions $G^j$ and gene encoding $E^j_i$ of each layer, depending on the depth $l$ of the specific architecture type. 
We set the genotype to be evolved $\mathbf{s}_j$ to \{$\beta_1,\beta_2,G^j$\}, where $\beta_1,\beta_2$ determine a truncated normal distribution with mean 0 and standard deviation $\beta_1$, truncated to the interval $[-\beta_2,\beta_2]$:

\begin{equation}
E^j_i = \begin{cases} 
-\beta_2 & \text{if } A^j_i < -\beta_2 \\
A^j_i & \text{if } -\beta_2 \leq A^j_i \leq \beta_2 \\
\beta_2 & \text{if } A^j_i > \beta_2 
\end{cases}
\end{equation}
where $A^j_i \sim \mathcal{N}(0, \beta_1^2)$. 

Since the parameters to be optimized are continuous, the Covariance Matrix Adaptation Evolution Strategy (CMA-ES) is a suitable evolutionary algorithm designed to solve complex optimization problems~\cite{hansen2001completely}.

The algorithm begins by initializing the mean vector $\mathbf{m}^{(0)} = [m_{\beta_1}, m_{\beta_2}, \mathbf{m}_O]$, the standard deviation vector $\sigma^{(0)} = [\sigma_{\beta_1}, \sigma_{\beta_2}, \sigma_O]$, and the covariance matrix $\mathbf{C}^{(0)} = \mathbf{I}$. 
At each iteration $t$, $\lambda$ candidate solutions are generated from a multivariate normal distribution: $\mathbf{s}_i = \mathbf{m}^{(t)} + \sigma^{(t)} \cdot \mathbf{B}^{(t)} \mathbf{D}^{(t)} \mathbf{z}_i$, where $\mathbf{z}_i \sim \mathcal{N}(\mathbf{0}, \mathbf{I})$, $\mathbf{B}^{(t)} \mathbf{D}^{(t)}$ is the decomposition of the covariance matrix, which is used to control the search direction and range. Based on the fitness $f_i$, the top $\mu$ samples are selected. The mean vector is updated as $\mathbf{m}^{(t+1)} = \sum_{i=1}^{\mu} w_i \mathbf{s}_{(i)}$ and $w_i$ are set according to fitness. The covariance matrix is updated using the evolution paths: 
\begin{equation}
\begin{aligned}
\mathbf{p}_\sigma^{(t+1)} &= (1 - c_\sigma) \mathbf{p}_\sigma^{(t)} + \sqrt{c_\sigma (2 - c_\sigma) \mu_\text{eff}} \mathbf{C}^{(t)^{-\frac{1}{2}}} (\mathbf{m}^{(t+1)} - \mathbf{m}^{(t)}) \\
\mathbf{p}_c^{(t+1)} &= (1 - c_c) \mathbf{p}_c^{(t)} + h_\sigma \sqrt{c_c (2 - c_c) \mu_\text{eff}} (\mathbf{m}^{(t+1)} - \mathbf{m}^{(t)})
\end{aligned}
\end{equation}

The covariance matrix itself is then updated: 
\begin{equation}
\begin{aligned}
\mathbf{C}^{(t+1)} &= (1 - c_1 - c_\mu) \mathbf{C}^{(t)} + c_1 \mathbf{p}_c^{(t+1)}\mathbf{p}_c^{(t+1)^T} \\
&+ c_\mu \sum_{i=1}^{\mu} w_i (\mathbf{s}_{(i)} - \mathbf{m}^{(t)}) (\mathbf{s}_{(i)} - \mathbf{m}^{(t)})^T
\end{aligned}
\end{equation}
Finally, the global step size is updated as: 
\begin{equation}
\sigma^{(t+1)} = \sigma^{(t)} \exp\left(\frac{c_\sigma}{d_\sigma} \left( \frac{\|\mathbf{p}_\sigma^{(t+1)}\|}{E\|\mathcal{N}(0, \mathbf{I})\|} - 1 \right)\right)
\end{equation}

When evaluating multiple solutions, directly using the training loss as the fitness may either be time-consuming or inaccurate (depending on the specific number of epochs). 
Therefore, we develop a spatial-temporal dynamical fitness function to get as close to the actual performance of each solution as possible within a limited time and computational cost.

Firstly, we propose a \textbf{Temporal Difference Regularization} that can apply different strengths of neuronal output at different stages of evolution.
In the early stages of evolution, a strong regularization:
\begin{equation}
r_1 = \sum_{t_s=0}^{T_s-1} \| Y^{(t_s+1)} - Y^{(t_s)} \|_F^2
\end{equation}
is used to select solutions where the output changes significantly between consecutive time steps, allowing the individual to learn and adapt to complex data features more freely and indicating more pronounced dynamic changes in the internal state, thereby encouraging the population to generate more innovative solutions.

For the gene interaction matrix $G$, we design a \textbf{Spatial Entropy Regularization} which adopt information entropy~\cite{jaynes1957information} to measure the complexity.
High values indicate that the gene network is able to capture and process complex input patterns. 
Normalize the elements of matrix $G$ so that their sum is 1, transforming it into a probability distribution. The formula for the normalized matrix $P$ is:
\begin{equation}
P_{ij}=\frac{G_{ij}}{\sum_{i=1}^{g} \sum_{j=1}^{g} G_{ij}}
\end{equation}
The information entropy regularization term is defined as:
\begin{equation}
r_2 = -\sum_{i=1}^{g} \sum_{j=1}^{g} P_{ij} \log(P_{ij})
\end{equation}

As the number of generations $t$ of evolution increases, the influence of the two regularization terms on fitness gradually decreases, and the evolutionary goal gradually becomes mainly based on the classification loss $L$ within $n_{eval}$ epochs of the solution.

\begin{equation}
f_i = L - e^{\lambda_1t} r_1  - e^{\lambda_2t} r_2
\end{equation}
where $\lambda_1$ and $\lambda_2$ are constants. The whole optimization process is shown in Algorithm.~\ref{alg:evo}.

\begin{figure}[htp]

  \renewcommand{\algorithmicrequire}{\textbf{Input:}}
  \renewcommand{\algorithmicensure}{\textbf{Output:}}
  \begin{algorithm}[H]

    \caption{The procedure of GEE algorithm.}
    \label{alg:evo}

    \begin{algorithmic}[1]
        \REQUIRE Number of generations $T_g$, Population size $n_{pop}$, Data $D$;
        \ENSURE Best individual model $Model$;
        \STATE Initialize population with $n_{pop}$ individuals;
         \WHILE {$t<T_g$}
            \STATE Generate $\lambda$ candidate solutions $s_1,s_2,...,s_\lambda$, $s_j$ = \{$\beta^j_1,\beta^j_2, G^j\}$, through mean vector $\mathbf{m}^{(t)}$, covariance matrix $\mathbf{C}^{(t)}$ and global step size $\sigma^{(t)}$;
            \STATE Initial wiring rule $E^j_1, E^j_2... E^j_l = Sample (\beta^j_1,\beta^j_2)$;
            \STATE Converting indirect encoding to computational model $Phenotype(s_j)$ = Encoding ($E^j_1, E^j_2... E^j_l,G^j$);
            \STATE Input $D$, apply temporal difference and spatial entropy regularization to calculate $f_i$ for each $Phenotype$;
            \STATE Select top $\mu$ samples based on fitness;
            \STATE Update $\mathbf{m}^{(t+1)}$, $\mathbf{C}^{(t+1)}$ and  $\sigma^{(t+1)}$;
            \STATE Calculate $f_i$ and select the best solution $s^* =\{\beta_1^*,\beta_2^*,G^*\}$;

        \ENDWHILE
        \STATE $Model$ = Encoding ($E^*_1, E^*_2... E^*_l,G^*$);
        \STATE {Optimize $\left \{ E^*_1, E^*_2... E^*_l,G^*\right \} $ with surrogate gradients;}
        \STATE {$Accuracy$ = Test($Model$)};
    \end{algorithmic}
  \end{algorithm}
\end{figure}

\begin{table*}[htbp]
\centering
  \large
  \resizebox{5.5in}{!}{%

  \setlength\tabcolsep{14pt} 
  \renewcommand{\arraystretch}{1.6} 
\begin{tabular}{ccccc}
\toprule
\textbf{Methods} & \textbf{Architecture} & \textbf{Param (M)}  & \textbf{Time Step} & \textbf{Accuracy (\%)} \\ 
\midrule

\multirow{1}{*}{Hybrid training \cite{rathi2020enabling}}
& ResNet-34 & 21.79 & 250 & 61.48 \\ 
\multirow{2}{*}{TET \cite{dengtemporal}}

 & Spiking-ResNet-34 & 21.79   & 6 & 64.79 \\ 
 & SEW-ResNet-34 & 21.79   & 4 & 68.00 \\

\multirow{2}{*}{SEW ResNet \cite{fang2021deep}}
 & SEW-ResNet-101 & 44.55   & 4 & 68.76 \\  
 & SEW-ResNet-152 & 60.19   & 4 & 69.26 \\ 

\multirow{2}{*}{GAC-SNN \cite{qiu2024gated}}
 & MS-ResNet-18 & 11.82    & 4 & 65.14 \\  
 & MS-ResNet-34 & 21.93  & 4&  70.42\\  

\multirow{2}{*}{Spiking ResNet \cite{hu2021spiking}} 
& ResNet-34 & 21.79  & 350 & 71.61 \\ 
 & ResNet-50 & 25.56   & 350 & 72.75 \\ 

\multirow{3}{*}{Spikformer \cite{zhouspikformer}} 
& Spikformer-8-384 & 16.81  & 4 & 70.24 \\
 & Spikformer-6-512 & 23.37   & 4 & 72.46 \\  
 & Spikformer-8-768 & 66.34   & 4 & 74.81 \\ 
\midrule
\multirow{1}{*}{\textbf{Our Method}} 
& ResNet-34 & 10.75   & 4 & 71.82 \\

 \bottomrule

\end{tabular}}
\caption{Comparison of various methods on ImageNet dataset. Param represents the number of parameters.}
\label{img}

\end{table*}

\section{Experiments}

In this section, we verify the effectiveness of the proposed method on CIFAR10~\cite{krizhevsky2009learning}, CIFAR100~\cite{xu2015empirical} and ImageNet~\cite{deng2009imagenet} datasets. To be fair, we apply GEE on multiple commonly used network architectures VGGNet~\cite{rathi2020enabling}, ResNet~\cite{zheng2021going}, CIFARNet~\cite{wu2018spatio}, and conduct comprehensive comparisons with various models to demonstrate the high efficiency, scalability, and robustness of GEE.

\subsection{Comparative Results}

Through genetic re-encoding and evolution, the proposed model can capture complex interactions between input and output features more effectively.
We compare the number of parameters and accuracy of the evolved candidate solutions on ImageNet and CIFAR10/CIFAR100, as shown in Table~\ref{img} and Table~\ref{cif} respectively. 
In ImageNet, in addition to parameter size, timesteps and accuracy, we also calculate the theoretical energy consumption as in previous works~\cite{rathi2023diet,zhouspikformer,yao2024spike,qiu2024gated}:
\begin{equation}
\begin{aligned}
E &=E_{MAC} \cdot FL_{\text {conv }}^1+E_{A C} \cdot T \cdot SOPs\\
SOPs &= \sum_{i_c=2}^{I_c} F L_{\text {conv }}^{i_c} \cdot f r^{i_c}+\sum_{i_f=1}^{I_f} F L_{fc}^{i_f} \cdot fr^{i_f}
\end{aligned}
\end{equation}
where $T$ is the time step, $fr$ is the firing rate of each layer, $I_c$ and $I_f$ are the number of convolutional layers and fully connected layers, respectively. Following the same assumptions of previous work, we set $E_{MAC} = 4.6 pJ$ and $E_{AC} = 0.9 pJ$. 
Although the proposed encoding scheme will increase the floating point operations per second, the sparse emission spikes make the energy consumption 43.24mJ lower than 59.295mJ of the model \cite{hu2021spiking} that also based on ResNet34. As can be seen from Table~\ref{img}, GEE shows a significant advantage in the number of parameters compared to other methods.
When the number of parameters is similar, GEE's accuracy is 6.68\% higher than GAC-SNN~\cite{qiu2024gated}. When the accuracy is similar, GEE compresses the number of parameters by 50\%. Compared with the transformer-based structure Spikformer~\cite{zhouspikformer}, GEE compresses about 40\% of parameters with an accuracy advantage of 1.56\%.


On both CIFAR10 and CIFAR100, GEE achieves high classification performance with very few parameters and the least time steps, whether based on CIFARNet, ResNet or VGGNet.
Compared with TSSL-BP~\cite{zhang2020temporal} which is also based on CIFARNet, GEE achieves a 2.06\% improvement on CIFAR10 with about 20\% of the parameters.
By increasing the number of genes, GEE achieves a further improvement of 2.32\% and 2.08\% on CIFAR10 and CIFAR100 respectively with about 7.5M parameters. Therefore, on CIFAR10, the accuracy is improved by 4.38\% in total, while compressing parameters by about 60\%.
Based on the VGG-16 architecture, GEE achieves 0.15\% improvements over the conversion-based method~\cite{hao2023reducing} with about 80\% fewer parameters on CIFAR10.
We conduct validations based on both ResNet18 and ResNet19.
Compared with other methods with the same architecture, GEE-ResNet19 only requires 34\% of the parameters, and improves performance by 1.5\%.
With only 40\% of the parameters of the transformer-based model Spikformer~\cite{zhouspikformer}, GEE-ResNet18 achieves a 1.66\% improvement in accuracy.
Compared with transformer-based models with similar parameters, GEE-ResNet19 improves by 1.61\% and 1.0\% on CIFAR10 and CIFAR100, respectively.

\begin{table*}[htbp]
\centering
  \large
  \resizebox{6in}{!}{%

  \setlength\tabcolsep{14pt} 
  \renewcommand{\arraystretch}{1.6} 
\begin{tabular}{cccccc}
\toprule
\textbf{Methods} & \textbf{Architecture} & \textbf{Param (M)} & \textbf{Timestep} & \textbf{CIFAR10(\%)} & \textbf{CIFAR100(\%)}  \\ \midrule
STBP \cite{wu2018spatio} & CIFARNet & 17.54 & 12 & 89.83 & - \\
STBP NeuNorm \cite{wu2019direct} & CIFARNet & 17.54 & 12 & 90.53 & - \\
TSSL-BP \cite{zhang2020temporal} & CIFARNet & 17.54 & 5 & 91.41 & - \\
\cmidrule{2-6}
\multirow{2}{*}{\textbf{Our Method}} 
 & CIFARNet & 3.72/3.81 & 4 & 93.47 & 72.97\\ 
 & CIFARNet & 7.44/7.53 & 4 & 95.79 &75.05\\ 
\midrule

Hybrid training \cite{rathi2020enabling} & VGG-11 & 9.27 & 125 & 92.22 & 67.87 \\

ANN2SNN~\cite{hao2023reducing} & VGG-16 & 33.6 & 32 & 95.42 & 77.34 \\
\cmidrule{2-6}
\multirow{2}{*}{\textbf{Our Method}} 
& VGG-16 & 2.51/2.70 & 4 & 94.47 & 75.02\\ 
& VGG-16 & 6.89/7.26 & 4 & 95.57 & 76.94\\ 
\midrule

Diet-SNN \cite{rathi2023diet} & ResNet-20 & 0.27 & 10/5 & 92.54 & 64.07 \\
STBP-tdBN \cite{zheng2021going} & ResNet-19 & 12.63 & 4 & 92.92 & 70.86 \\
TET \cite{dengtemporal} & ResNet-19 & 12.63 & 4 & 94.44 & 74.47 \\ 




\multirow{2}{*}{Spikformer~\cite{zhouspikformer}} & Spikformer-4-256 & 4.15 & 4 & 93.94 & 75.96 \\
 & Spikformer-4-384 & 9.32 & 4 & 95.19 & 77.86 \\ 
\multirow{1}{*}{GAC-SNN \cite{qiu2024gated}}
 & MS-ResNet-18 & 12.63/12.67   & 4 & 96.24 & 79.83 \\  
\cmidrule{2-6}

 \multirow{4}{*}{\textbf{Our Method}}
& ResNet-18 & 3.66/3.68 & 4 & 95.60 & 77.32 \\ 
& ResNet-18 & 8.72/8.75 & 4 & 96.26 & 78.91\\ 
& ResNet-19 & 4.35/4.38 & 4 & 95.94 & 77.56\\ 
& ResNet-19 & 8.41/8.44 & 4 & 96.80& 78.86\\ 

\bottomrule
\end{tabular}}
\caption{Comparison of different architectures on CIFAR10 and CIFAR100 datasets.}
\label{cif}

\end{table*}

\begin{table}[hpt]
\centering
  \resizebox{3.2 in}{!}{
\begin{tabular}{lllll}
\toprule
Architectures &Models &Acc (\%) & Spikes(K)\\
\midrule
\multirow{2}{*}{ CIFARNet} 
& \cite{wu2019direct} & 87.80 & $1298 $ \\
& \cite{fang2021incorporating} & 93.15 & $507 $ \\
&GEE-CIFARNet & 95.79 & $14$\\
\midrule
\multirow{3}{*}{ ResNet} 
&  ResNet11~\cite{lee2020enabling} & 90.24 & $ 1530 $\\
& ResNet19~\cite{zheng2021going} & 93.07 & $1246$ \\
&GEE-ResNet19 & 96.80 & $3$ \\
\bottomrule
\end{tabular}}
\caption{Comparison of the number of spikes on CIFAR10.}
\label{spikes}
\end{table}
We compare the fired spikes with other models on CIFAR10 as another way to evaluate energy consumption.
The results show that the amount of spikes on both CIFARNet and ResNet based GEE models is much smaller than that of other methods with the same architecture, as shown in Table.~\ref{spikes}. 
On CIFARNet, GEE achieves a 2.64\% improvement in accuracy with only 3\% spikes of other method with the same architecture~\cite{fang2021incorporating}.
On ResNet-19, GEE achieves 96.8\% accuracy with only 3K spikes, while STBP-tdBN~\cite{zheng2021going}, which is also based on ResNet-19, requires 400 times the number of spikes.
GEE achieves sparser spikes and more efficient paradigms, broadening the application of brain-inspired computation.

\subsection{Robustness}

\begin{figure}[htp]
\centering
\includegraphics[width=8.5cm]{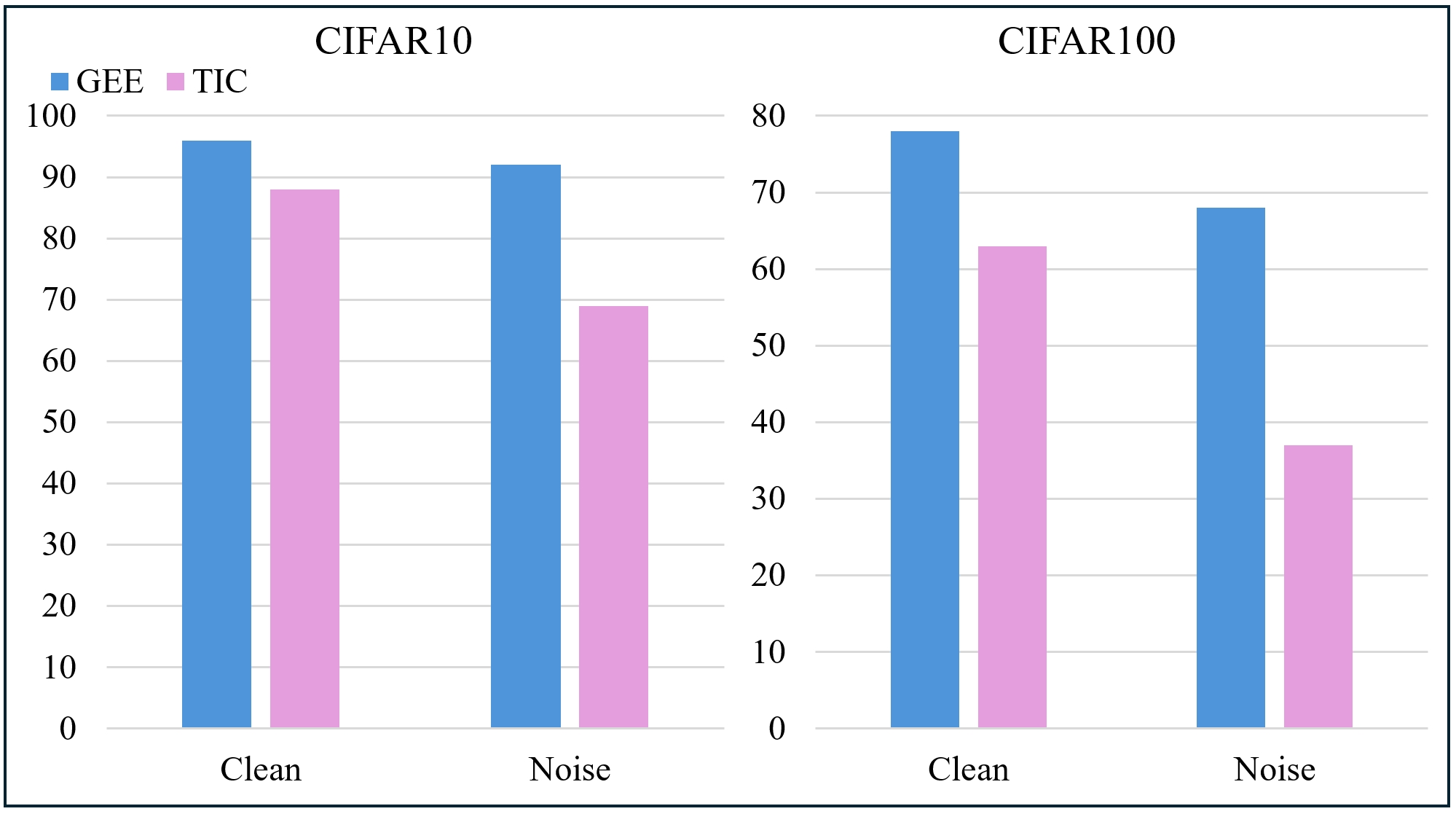}
\caption{Robustness verification on CIFAR10 and CIFAR100.}
\label{ro}
\end{figure}

To verify the robustness of GEE, we compare it with TIC~\cite{kim2023exploring} and adopt the same method of introducing Gaussian noise (add Gaussian noise whose $L_2$ norm is set to 50\% of the
norm of the given input) as shown in Fig.~\ref{ro}. 
The results show that in a noise-free scene, GEE is 2\% and 3\% more accurate than TIC, on CIFAR10 and CIFAR100 respectively. 
Under the influence of noise, the accuracy of TIC decreased by 23\% and 31\%, respectively, while the accuracy of GEE are 92.2\% and 68.7\%, only dropping by 6\% and 14\%.
Whether in a noise-free scene, a noisy scene, or the influence of noise, GEE performs significantly better than TIC, reflecting its noise resistance and the robustness brought to it by evolution.

\subsection{Ablation Study}

\begin{figure}[htp]
\centering
\includegraphics[width=7cm]{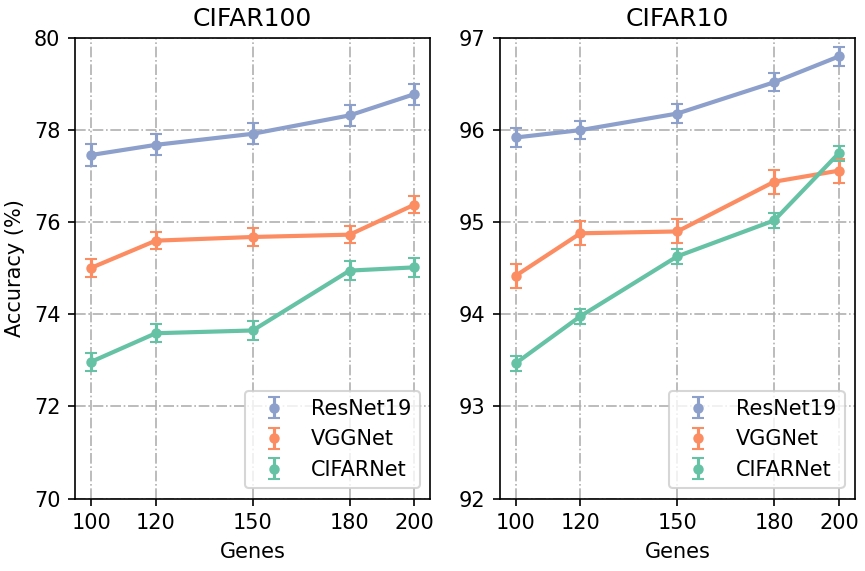}
\caption{Scalability of different architectures on CIFAR10 and CIFAR100.}
\label{f1}
\end{figure}

\textbf{Gene Scale.}
In the proposed gene-encoded SNN, the number of genes affects the encoding ability of the network.
In theory, the more genes there are, the more complex information patterns can be captured. Therefore, we verify it on CIFAR10 and CIFAR100, as shown in Fig.~\ref{f1}. As the number of genes increases, the proposed genetically encoded method (in VGGNet, CIFARNet, and ResNet) has been empirically proven to enhance classification performance on CIFAR10 and CIFAR100. 
As the intermediate representations become richer, allowing the model to express more intricate feature combinations and improve classification performance, demonstrating excellent scalability and flexibility.
By adjusting $g$, the model can be extended to improve performance, as shown in Table~\ref{cif}. Even if the number of genes has been compressed to a very small number, only about 30\% of the parameters of the model with the same architecture~\cite{dengtemporal}, GEE still achieves 2\% and 1.6\% improvement on the two datasets.
Different genetic scales show the same performance advantages in the model,
which shows that it is able to adapt to different complexity, thus highlighting the effectiveness and practicality.

\noindent \textbf{Regularization Term.}
In order to test the proposed regularization terms of STE, we fix the number of genes to 150 and construct four ablation models: Random model (the initial $X_i, O$ follows $X_i, O\sim \mathcal{N}(0, 1)$ without evolution), Baseline ($f_i = L$), Baseline + $r_1$ ($f_i = L+r_1$), Baseline + $r_2$ ($f_i = L + r_2$) and STE ($f_i = L + r_1 + r_2$). The comparison results are shown in Table~\ref{abl}.
The results are the average of multiple trainings and the number of training epochs is fixed to 200.
It can be seen that on ResNet, the random model without evolution has a much lower average performance and a larger variance than other models due to the gradient vanishing problem.
Both the regularization term $r_1$ and $r_2$ drive the efficiency of evolution (compared to the evaluation based only on training loss $L$) on both ResNet and VGGNet.
The solution evolved by STE is optimal, surpassing single regularization models and improving accuracy by 1.94\% and 2.24\% on CIFAR10 and CIFAR100 compared to Basline.
On VGGNet, no gradient vanishing is observed and STE improves the accuracy of the solution by 1.77\% and 1.80\% compared to the randomly initialized model. Both $r_1$ and $r_2$ drive the improvement of the solution quality, demonstrating the effectiveness of the spatio-temporal evolution framework.

\begin{table}[ht]
    \centering  
    \resizebox{3.3in}{!}{%

    \begin{tabular}{llll}
        \toprule
        \textbf{Architecture} & \textbf{Models} & \textbf{CIFAR10(\%) } &\textbf{CIFAR100(\%) } \\
        \midrule
        
        \multirow{5}{*}{ResNet-19} 
                    & Random & 54.96 $\pm$ 5.89& 35.94 $\pm$ 3.74\\
                    & Baseline & 94.12 $\pm$ 0.75& 75.66 $\pm$ 0.71 \\
                    & Baseline + $r_1$ & 95.47 $\pm$ 0.49& 76.80 $\pm$ 0.18\\
                    & Baseline + $r_2$ & 95.02 $\pm$ 0.66 & 77.10 $\pm$ 0.25\\
                    & \textbf{STE} & \textbf{96.16} $\pm$ 0.13 & \textbf{77.90} $\pm$ 0.06\\
                    
            \midrule

        \multirow{5}{*}{VGGNet}            
                    & Random & 93.63 $\pm$ 0.85& 73.89 $\pm$ 0.91 \\
                    & Baseline & 93.89 $\pm$ 0.94 & 74.95 $\pm$ 0.76 \\
                    & Baseline + $r_1$ & 95.18 $\pm$ 0.37& 75.44 $\pm$ 0.27 \\
                    & Baseline + $r_2$ & 95.21 $\pm$ 0.20  & 75.58 $\pm$ 0.33\\
                    & \textbf{STE} & \textbf{95.40} $\pm$ 0.12 & \textbf{75.69} $\pm$ 0.25\\

        \bottomrule
    \end{tabular}}
\caption{Performance of different mechanisms on CIFAR10 and CIFAR100.}
    \label{abl}
\end{table}

\subsection{Evolutionary Efficiency} 
\begin{figure}[htp]
\centering
\includegraphics[width=5.5cm]{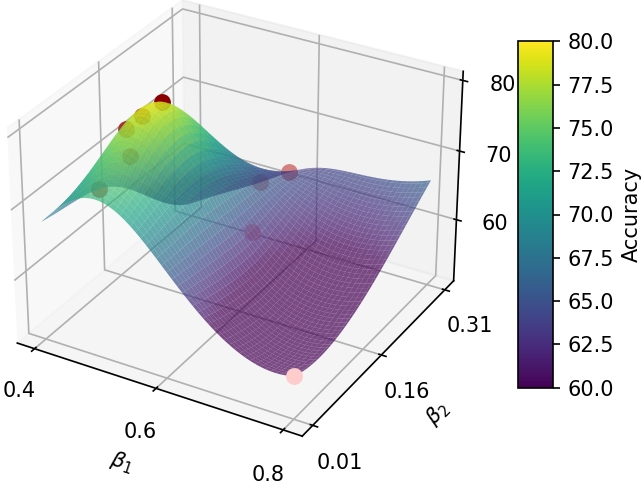}
\caption{Visualization of parameter evolution on CIFAR10.}
\label{heat}
\end{figure}

To further evaluate the efficiency of the proposed STE, we compared the time required for model Baseline (using only the training loss as the optimization objective) and STE within 5 generations. We extend the training epochs of Baseline to 10, while STE remains unchanged.
The experiments are conducted on a system equipped with seven NVIDIA A100 GPUs and an AMD EPYC 7H12 64-Core Processor, running on a Linux environment with Ubuntu 20.04.
The results show that after extending the training time, Baseline improves the accuracy of the evolved solution by 1.2\% on the original basis, but still lags behind STE by 0.84\%. 
In addition, the evolution time required for STE is about five gpu hours, while Baseline requires 25 gpu hours, which is five times that of STE.
We visualize the evolution of $\beta_1$ and $\beta_2$ as shown in Fig.~\ref{heat}. We sample some candidates and test their accuracy on CIFAR10. It can be seen that the light-colored dots represent the poor performance of the initial individuals. As the evolution progresses, the classification accuracy of the individuals gradually improves, which illustrates the effectiveness of the spatio-temporal evolutionary algorithm.

\section{Conclusion}

This study introduces a genetically encoded evolutionary  spiking neural network that leverages a compact, gene-scaled neuronal coding scheme to effectively reduce the parameter count and computational demands typical of conventional SNNs.
By embedding a spatio-temporal evolution framework, the model not only improves the quality of the solution but also significantly reduces energy consumption. The dynamic regularization mechanism improves the solution's ability to effectively evolve and stabilize functional neuronal patterns.
Empirical validation on CIFAR10, CIFAR100, and ImageNet shows that GEE not only reduces computational overhead, but also maintains efficiency compared to traditional methods, demonstrating the potential of brain-inspired architectures to achieve high efficiency and low power consumption.

\bibliography{IEEEabrv,bibfile}

\end{document}